\documentclass[runningheads]{llncs}
\usepackage{graphicx}
\usepackage{hyperref}
\hypersetup{
    colorlinks=true,
    linkcolor=blue,
    filecolor=blue,      
    urlcolor=blue,
}
\usepackage{amsmath, amssymb}
\begin{document}
\title{Estimating Direct and Indirect Causal Effects of Spatiotemporal Interventions in Presence of Spatial Interference}
%
%\titlerunning{Abbreviated paper title}
% If the paper title is too long for the running head, you can set
% an abbreviated paper title here
%
\author{Sahara Ali\inst{1,2} \and
Omar Faruque\inst{1} \and
Jianwu Wang\inst{1}}
\authorrunning{Ali et al.}
\titlerunning{Estimating Direct and Indirect Causal Effects}
% First names are abbreviated in the running head.
% If there are more than two authors, 'et al.' is used.
%
\institute{University of Maryland Baltimore County, Baltimore, MD, United States \\
\email{\{sali9,omarf1,jianwu\}@umbc.edu} \\
\url{https://bdal.umbc.edu}  \and
University of North Texas, Denton, TX, United States}
\maketitle              % typeset the header of the contribution
\begin{abstract}
Spatial interference (SI) occurs when the treatment at one location affects the outcomes at other locations. Accounting for spatial interference in spatiotemporal settings poses further challenges as interference violates the stable unit treatment value assumption, making it infeasible for standard causal inference methods to quantify the effects of time-varying treatment at spatially varying outcomes. In this paper, we first formalize the concept of spatial interference in the case of time-varying treatment assignments by extending the potential outcome framework under the assumption of no unmeasured confounding. We then propose our deep learning based potential outcome model for spatiotemporal causal inference. We utilize latent factor modeling to reduce the bias due to time-varying confounding while leveraging the power of U-Net architecture to capture global and local spatial interference in data over time. Our causal estimators are an extension of average treatment effect (ATE) for estimating direct (DATE) and indirect effects (IATE) of spatial interference on treated and untreated data. Being the first of its kind deep learning based spatiotemporal causal inference technique, our approach shows advantages over several baseline methods based on the experiment results on two synthetic datasets, with and without spatial interference. Our results on real-world climate dataset also align with domain knowledge, further demonstrating the effectiveness of our proposed method. 
\end{abstract}
\section{Introduction}
\begin{figure}[ht!]
\centering
  \includegraphics[width=\textwidth]{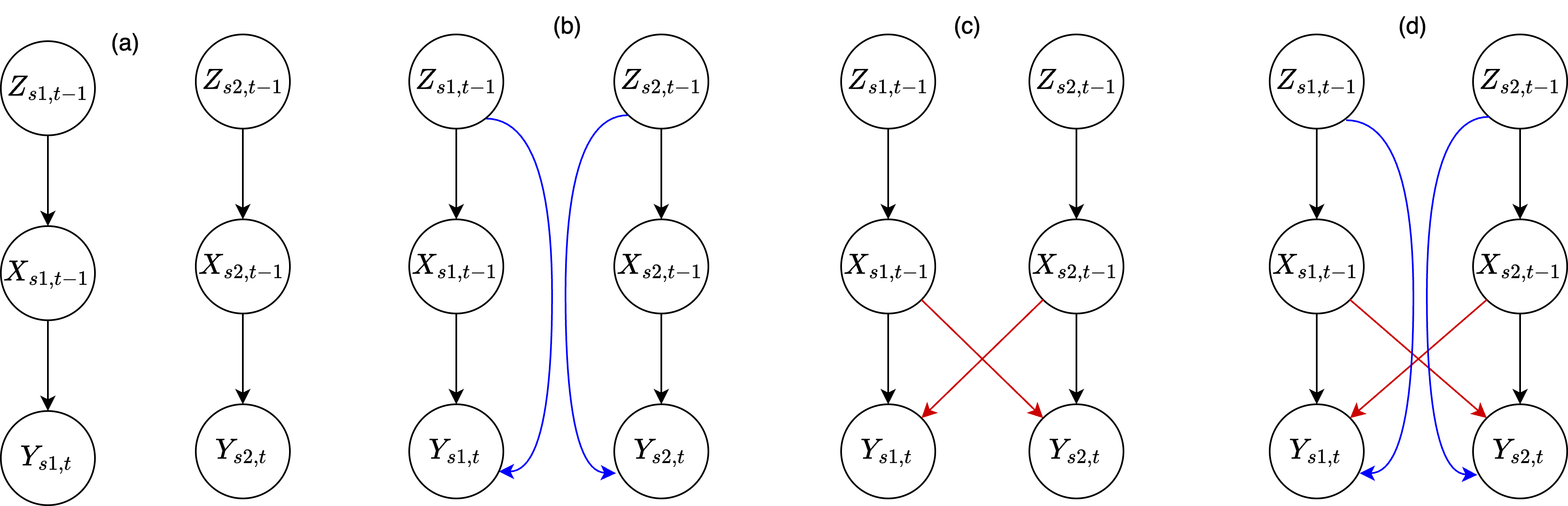}
  \caption{Different scenarios of causation (black), confounding (blue) and interference (red) in spatiotemporal data. (a) No confounding, no interference, only temporal causation; (b) No interference, only temporal confounding and temporal causation; (c) No temporal confounding, only spatial interference and temporal causation; (d) Temporal confounding, spatial interference, temporal causation.}
  \label{fig:conf}
\end{figure}

Quantifying the effects of an entity, process or state, referred as the cause, on another entity, process or state, referred as the outcome, is an active research area with wide applications in epidemiology, economics, political and environmental science~\cite{reich2021review,yao2021survey}. 
In many real-world scenarios, the effect of a treatment or intervention is not static but evolves dynamically over space and time, presenting a challenge for accurate estimation of time-varying treatment effects under spatial interference. The concept of spatial interference stems from spatial statistics where interventions or events at one location not only impact outcomes at that specific location but also propagate to neighboring or distant locations~\cite{ripley1988statistical}. This spatial dependency violates the assumptions of independence and stable unit treatment value (SUTVA)~\cite{imbens2015causal}, underlying many conventional causal inference techniques,  which restricts every unit in the data to have treatment applied to only its own outcome. When spatial interference is present, the treatment at one unit spills over and influences the outcome of neighboring units. Oftentimes, spatial interference is confused with spatial confounding. Confounding occurs when the past values of some covariates influence the current values of treatment and outcome leading to spurious correlations and biases in the outcome values. We refer to such covariates as confounders~\cite{pearl2009simpson}. In Figure~\ref{fig:conf}, we illustrate four possible spatial and temporal interactions in data at two locations $s1$ and $s2$ over timesteps $t$ and $t-1$. Here, $X$ is the treatment variable, $Z$ is a covariate, and $Y$ is the outcome. In this paper, we present the phenomenon in Figure ~\ref{fig:conf}d, i.e., how time-varying treatment affects potential outcome in the presence of time-varying confounding and spatial interference. Though the real-world observational data comprises intricate complexities of both spatial and temporal confounding, we limit our scope to observed temporal confounders, whereas identifying confoundedness in space or through unobserved confounders is beyond the scope of this paper. 

We rely our work on three major arguments. (1) Traditional linear models used for estimating causal effects rely heavily on strong parametric assumptions. Conversely, neural networks, being nearly non-parametric in nature, offer the advantage of capturing the diverse treatment effects observed across individual units while minimizing bias~\cite{koch2021deep}. (2) Even though causal effect estimation in temporal and spatial settings have been investigated previously (see Section~\ref{related-work}), there exists limited work that handles both of these tasks simultaneously. (3) We further argue propensity score based techniques are computationally expensive and unable to handle continuous time-varying treatments~\cite{bica2020estimating,ali2023quantifying}.
%In light of these arguments, we propose a spatiotemporal causal inference network to estimate the time-varying treatments on spatially interfering neighborhoods. 

\textbf{Our Contributions.} (1) We extend the potential outcome framework to spatiotemporal setting by introducing STCINet: a spatiotemporal causal inference network based on U-Net architecture with double attention to learn causal relations with spatially interfering treatments. (2) We propose an autoencoder based factor model to reduce the time-varying confounding effect of spatial data by adapting the factor model introduced in~\cite{bica2020time} and extending it to spatiotemporal domain. (3) We establish the case of spatial interference in time-varying data by evaluating our method on synthetic datasets based on diffusion phenomenon. (4) We establish a promising research direction for spatiotemporal causal inference in climate science by quantifying the direct and indirect effects of atmospheric processes on Arctic sea ice melt. Our implementation code can be accessed on  GitHub.
%\footnote{https://github.com/iharp-institute/causality-for-arctic-amplification/tree/main/stcinet}
\footnote{https://tinyurl.com/stcinet}

The rest of the paper is organized as follows. Section~\ref{sec2} enlists the causal assumptions and notations followed throughout this paper. Section~\ref{sec3} explains the overall architecture and individual modules of our proposed method. Section~\ref{sec4} mentions the data generation process, experimental configurations, evaluation methods and empirical results of our model on synthetic and real world data. Section~\ref{sec5} highlights the related work in causal inference. Lastly, we conclude our paper in Section~\ref{sec6} and mention some potential extensions of this work.

\section{Preliminaries}
\label{sec2}
\subsection{Notations and Definitions}
\label{definitions}
Given spatiotemporal data over $N \times M$ region and spanning over $T$ timesteps, $X^t = {X}^t_{i,j \in [N,M]}$ represents the treatment variable at timestep $t \in T$, $Z^t = {Z}^t_{i,j \in [N,M]}$ represents a set of time-varying covariates and $Y^t = {Y}^t_{i,j \in [N,M]}$ represents the outcome, such that at every location $\{X_{i,j}^t, Z_{i,j}^t, Y_{i,j}^t\}$ $\in$ $\mathbb{R}$. When intervened on the value of treatment $X$, the corresponding updated value for the same \((i,j)\) location is represented by $\hat{X}_{i,j}$, where $\hat{X}_{i,j} = update\_factor \times X_{i,j}$. At any given time $t$, $Y_{i,j}(\hat{X}_{i,j})$ is the potential outcome under intervened treatment $\hat{X}$, and $Y_{i,j}(X_{i,j})$ is the potential outcome under un-intervened treatment $X$ (also called placebo effect). 
Further, at any given timestep $t$, $\bar{X^t}$ represents the set of historic values of $X$ and $\bar{Z^t}$ represents the set of historic values of $Z$, such that $\bar{X}^{t} = (X^1, X^2, X^3,..., X^{t-1})$ and $\bar{Z}^{t} = (Z^1, Z^2, Z^3,..., Z^{t-1})$. 

\textbf{Outcome under no Spatial Interference (Figure~\ref{fig:conf}a).}
Assuming $Y^{t+l}$ to be dependent on $X^{t}$ and an unknown noise term $\varepsilon_y$, a simple case of data-generation process under no spatial interference and only time-varying lagged dependence can be given by $Y^{t+l}_{i,j} = \beta X^{t}_{i,j} + \varepsilon_y$ , where $X^{t}_{i,j} = \gamma Z^{t}_{i,j} + \varepsilon_x$. Here, $\beta$ and $\gamma$ represent the causal coefficients of $X$ and $Z$ and $l$ is the temporal lag.

\textbf{Outcome under Spatial Interference (Figure~\ref{fig:conf}c).}
\label{eq:2}
Next, we assume that the outcome $Y$ at every location is also influenced by the treatment applied at neighboring locations. The data-generation process will now comprise spatial interference and time-varying lagged dependence, given by $Y^{t+l}_{i,j} = \beta_1 X^{t}_{i,j} + \beta_2 \tilde{X}^{t} + \varepsilon_y $,
where, $\tilde{X}$ represents mean of $m$ neighborhood values of $X^t_{i,j}$ at any given timestep, such that $\tilde{X} = \mathbb{M}(X_{(i-m,j-m),...(i+m,j+m)})$, excluding $X^{t}_{i,j}$ itself in the mean computation. Here, $\beta_2$ represents the causal coefficient of $\tilde{X}$.

\textbf{Outcome under Spatial Interference and Temporal Confounding (Figure~\ref{fig:conf}d).}
\label{eq:3}
Here, we extend the data-generation process to settings where $Z$ is the confounder, i.e., both $X$ and $Y$ are dependent on $Z$, given by $Y^{t+l}_{i,j} = \beta_1 X^t_{i,j} + \beta_2 \tilde{X}^{t} + \beta_3 Z^t_{i,j} + \varepsilon_y$, and $X^{t}_{i,j} = \gamma Z^{t}_{i,j} + \varepsilon_x$.

Under the setting of spatial interference and temporal confounding, we present three different metrics for estimating the causal effects of $X$ on $Y$ in the presence of confounder $Z$; (1) in the treated sub-region, (2) in the untreated sub-region where spillover is observed and (3) finally in the overall spatial region of a given dataset.

\textbf{Definition 1: Direct (Spatial) Average Treatment Effect~\cite{tec2023weather2vec}.}
Given observed $X$ and intervened $\hat{X}$, referring to time-varying treatment values, at time-step $t$ and a spatial location $s \in \mathbb{S}$, such that $ \mathbb{S} \subset N \times M$, the direct treatment effect $\tau_{date}$ refers to the difference in potential outcomes $Y$ under $\hat{X}$ and $X$, observed directly on the treated location $s$. The direct average treatment effect $\tau_{date}$ is given by:
\begin{equation}
    \tau_{date} = \frac{\Sigma_{s\in\mathbb{S}} (Y_s(\hat{X}_s,Z) - Y_s(X_s,Z))}{|\mathbb{S}|}
    \label{eq:date}
\end{equation}

\textbf{Definition 2: Indirect (Spatial) Average Treatment Effect~\cite{lok2016defining}.}
The indirect spatial treatment effect $\tau_{iate}$ refers to the difference in potential outcomes under $\hat{X}$ and $X$, observed on the untreated region ${s'} \in \mathbb{S'}$, such that $ \mathbb{S'} \subset N \times M$ and $s' \neq s $, where $s$ refers to the treated region. The indirect average treatment effect $\tau_{iate}$ is given by:
\begin{equation}
    \tau_{iate} = \frac{\Sigma_{{s'}\in\mathbb{S}'} (Y_{s'}(\hat{X_s},Z) - Y_{s'}(X_s,Z))}{|\mathbb{S}'|}
    \label{eq:iate}
\end{equation}

\textbf{Definition 3: Lagged Average Treatment Effect~\cite{ali2023quantifying}.}
Combining the net effect of direct and indirect treatment over the entire region $\mathbb{N} = N \times M$, the overall average treatment effect $\tau_{late}$ at a temporal lag of $l$ is given by:
\begin{equation}
    \tau_{late} = \frac{\Sigma_{{k}\in\mathbb{N}} (Y^{t+l}_{k}(\hat{X^t_{k}},Z^t_{k}) - Y^{t+l}_{k}(X_{k},Z_{k}))}{|\mathbb{N}|}
    \label{eq:late}
\end{equation}

\subsection{Assumptions}
Extending our understanding developed in Section~\ref{eq:3}, we assume that the outcome is generated by treatment, covariates and noise, given by $Y = \digamma(X, Z) + \varepsilon_y$, where $\digamma$ is an unknown and non-linear function. Further, the following assumptions hold for the method and experiments proposed in the remaining paper:

\textbf{Assumption 1: No Unmeasured Confounding~\cite{hernan2006estimating}.}
We assume there is no unmeasured or unobserved confounding other than the confounding caused by observed covariates $Z$. Further, the confounding is only limited to temporal scale such that $Y^t_{i,j} = \digamma(X,Z^t_{i,j},\bar{Z}_{i,j})$ and $Y^t_{i,j} \neq \digamma(X,Z^t_{i',j'},\bar{Z}_{i',j'})$ where $(i,j) \neq(i',j')$.

\textbf{Assumption 2: Consistency~\cite{cole2008constructing}.}
If the historic values of treatment are $\bar{X}^{t} = \bar{x}^t$, then the potential outcome under the treatment is the
same as the observed (factual) outcome $Y(\bar{x}^t) = Y$.

\textbf{Assumption 3: SUNTVA.} We replace the stable unit treatment value assumption with its variant - Stable Unit Neighborhood Treatment Value Assumption (SUNTVA), introduced by~\cite{forastiere2021identification}, to accommodate spatial interference. Under SUNTVA, for each location $s$, there exist a neighborhood $\mathbb{M}_s$, such that the outcome is influenced by $X_s$ as well as by the neighborhood of treatment $X_{\mathbb{M}s}$, such that $Y_s$ = $Y_s(X_s, X_{\mathbb{M}s})$.

In addition, we assume positivity~\cite{rubin2005causal}, such that at each timestep $t$, each treatment has a non-zero probability of being assigned to the outcome in the treated region. Since we refer to a spatiotemporal setting, we relax the spatial subscript $(i,j)$ moving forward, to mention the time-varying aspect of data, implying $Y_{i,j}^t \xrightarrow{} Y^t$ unless specified otherwise.

\section{\underline{S}patio-\underline{T}emporal \underline{C}ausal \underline{I}nference \underline{Net}work (STCINet)}
\label{sec3}
\begin{figure*}[ht!]
\centering
\includegraphics[width=1.0\textwidth]{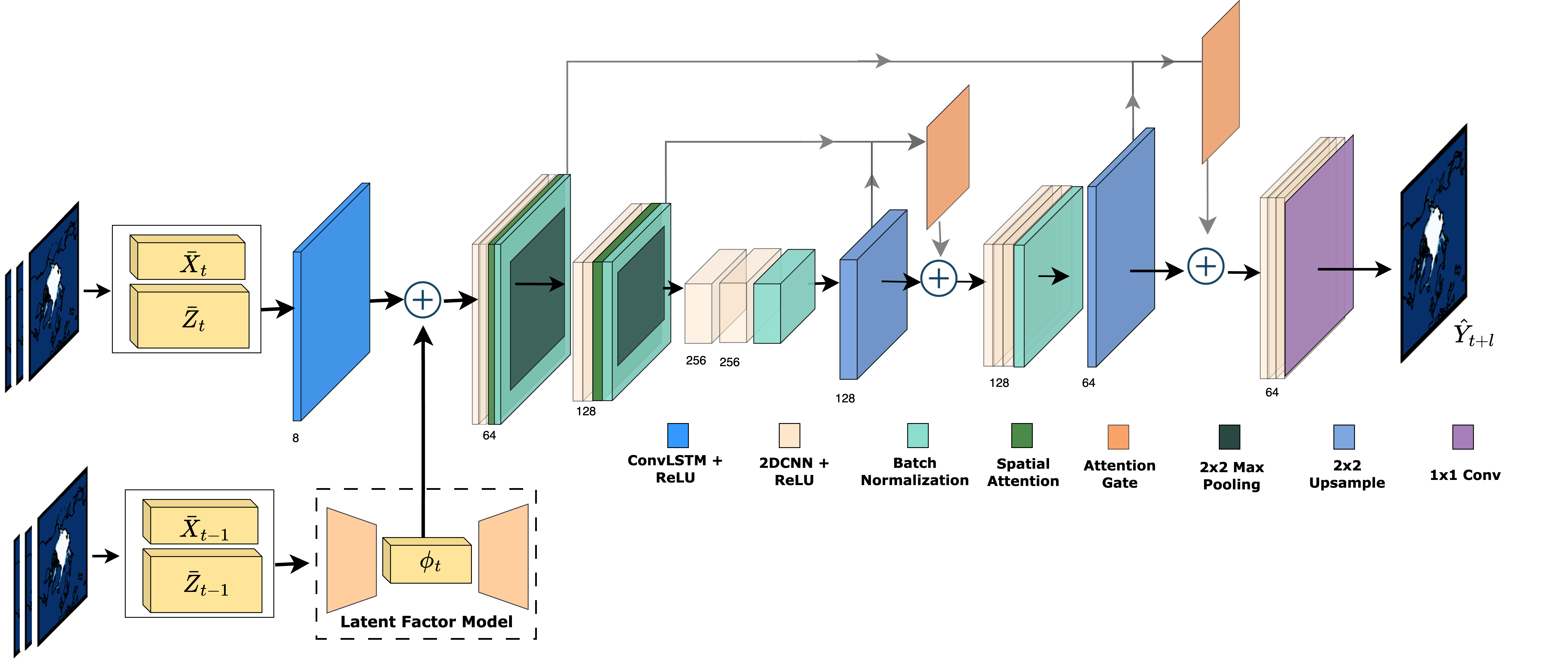}
  \caption{Overall architecture of proposed spatiotemporal causal inference network (STCINet).}
  \label{fig:arch}
\end{figure*}

Here, we present our proposed technique to perform causal inference under time-varying confounding and spatial interference.
%\subsection{Overall Pipeline}
The overall architecture is given in Figure~\ref{fig:arch}, where we first divide the spatiotemporal data into current data $(X^t,Z^t)$  and history data $(\bar{X}^{t-1},\bar{Z}^{t-1})$. The current data passes through a Convolutional Long Short Term Memory (ConvLSTM) layer, to learn the lagged representation of treatment and covariates, whereas the historic data is fed to the Latent Factor Model (LFM), through which the uncorrelated latent representations of treatment and covariates are learned, assuming the covariates also comprise some observed confounders. We employ the LFM technique as opposed to the propensity score weighting (IPTW)~\cite{robins2000marginal} owing to the promising results of factor models observed in recent literature~\cite{bica2020estimating,bica2020time}. The latent representation for $\phi^t$ is then combined with the output of ConvLSTM. This combined set of features is fed as the input to a U-Net based model. With the aid of attention gating, the custom U-Net model learns the local and global spatial variations in data, occurring in response to the treatment application, and finally, the U-Net predicts the future values of outcome $Y^{t+l}$ after a lag of $l$ timestep. At test time, the trained model is fed treated (intervened) and untreated (observed) values of treatment variable to estimate direct, indirect and overall causal effects using the metrics of $\tau_{date}, \tau_{iate}$ and $\tau_{late}$. 
Below, we explain the functionality of each of these modules in detail.

\subsection{Latent Factor Model for Temporal Confounding}
\begin{figure}[h]
  \centering
  \includegraphics[width=\linewidth]{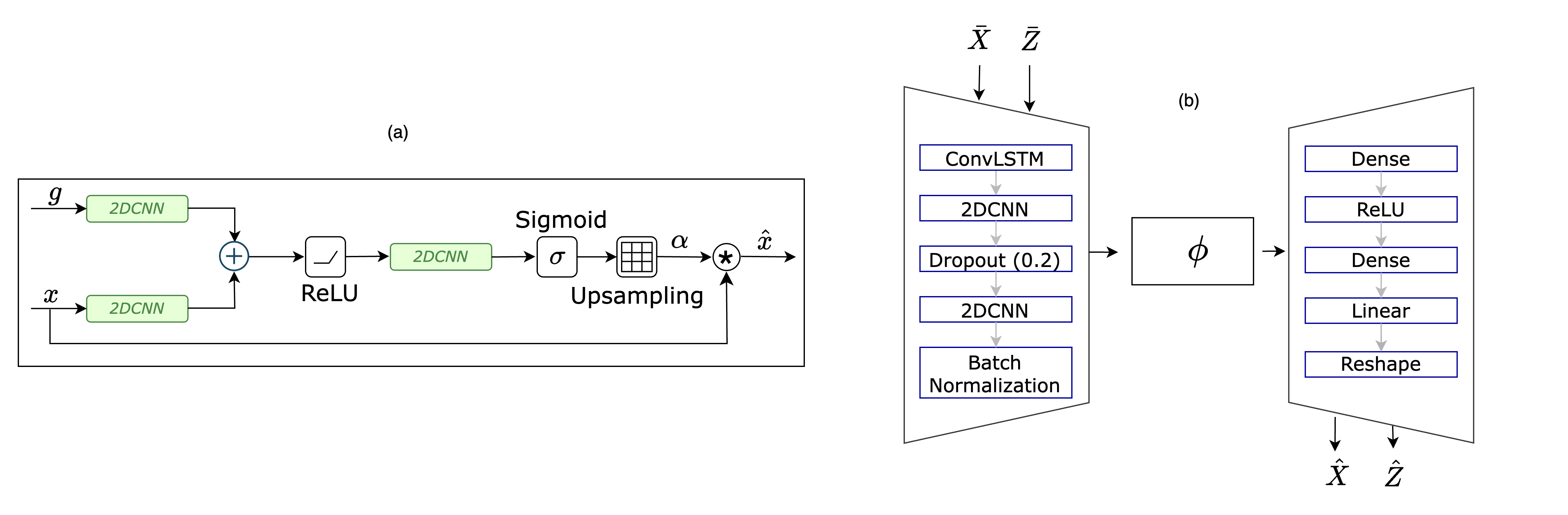}
  \caption{Sub-modules of STCINet: (a) Attention gating mechanism to identify patterns of spatial interference, (b) Latent factor model for deconfounding covariate and treatment history.}
\label{fig:lfm}
\end{figure}

Latent factor models are designed to uncover the hidden structure or underlying factors that influence the observed data~\cite{ghojogh2021factor}. Latent variables are not directly measured or observed but are inferred from the observed data patterns. To observe the sole influence of current treatment on the future outcome, in the presence of time-varying confounders, we propose a latent factor model (LFM), inspired by~\cite{bica2020time} that learns the distribution of treatment and covariates over time and de-correlates the entangled relationship so the outcome becomes independent of the confoundedness, i.e., $Y^t \perp \bar{Z}^{t-1} | X^t$.  
Our implementation of the latent factor model is given in Figure~\ref{fig:lfm}b, where we design an autoencoder based model that takes in historic spatiotemporal values of treatment and covariates just before the timestep where treatment is applied, i.e., $t-1$. 

The encoder part of the LFM comprise of one ConvLSTM layer, one 2D convolution (CNN) layer followed by dropout and the second 2D CNN layer followed by batch normalization. While the purpose of ConvLSTM and CNN layers is to learn the time-varying spatial features of historic data, dropout and normalization is applied to reduce overfitting and improve the generalization of the encoder. The output of the encoder is given by $\phi^t = \text{Enc}((\bar{X}^{t-1},\bar{Z}^{t-1}))$.

The decoder comprise of one fully connected layer with ReLU activation and the second fully connected layer with linear activation. Finally, we reshape the outcome back the dimensions of $X$ and $Z$. We use the Mean Squared Error (MSE) as the reconstruction loss for the decoder. The reconstructed output of the decoder is given by $(\hat{X}^{t-1},\hat{Z}^{t-1}) = \text{Dec}(\phi^t)$.
\subsection{Double Attention Mechanism}
Attention mechanism allows the model to selectively focus on different parts of the input by assigning attention weights to each element in the input based on its contribution in predicting the outcome~\cite{vaswani2017attention}. The selective focus helps the model to ignore less important parts of the input thereby improving the model's predictive performance. Attention mechanism has previously shown promising contribution in causal discovery problems~\cite{nauta2019causal}, whereas employing attention in U-Net has shown to increase model's sensitivity to local and global variations~\cite{oktay2018attention}. In our model, we apply attention at two stages; (i) in the STCINet downsampling block and (ii) in the STCINet upsampling block. We refer to the overall role of attention as double attention mechanism for STCINet.

\textbf{Spatial Attention.} Spatial attention is added to the downsampling part of STCINet, where our goal is to enable the model to selectively attend to specific spatial regions that are most affected by treatment assignment. This is done by performing max pooling and average pooling separately on the output of downsampling blocks. Both the pooling outcomes are concatenated and passed through sigmoid activation to get per pixel attention weights. These weights are applied to the downsampling block's output by element-wise tensor multiplication.

\textbf{Attention Gating.} In the STCINet upsampling block, we incorporate the attention-gating (AG) mechanism introduced by~\cite{jetley2018learn} for our spatial interference task. As shown in Figure~\ref{fig:lfm}a, our AG module takes in two inputs, $x$ and $g$, referring to the input and the gating signal, respectively. The key idea is to assign weights to local regions within $x$ that align with the location of global features in gating signal $g$. Here, spatial regions are selected by analysing the contextual information provided by the gating signal $g$ which is collected from a coarser scale. In case of STCINet, $g$ represent the skip-connection from the downsampling blocks whereas $x$ represents the upsampled output from the previous upsampling layer. Both the inputs first pass through 2D CNN layers, to align their depth (filters) and dimensions (height, width). We then perform element-wise addition of the transformed inputs, followed by ReLU activation and $1\times1$ 2D convolution, to reduce the number of trainable parameters in gating operation. Finally, we use sigmoid activation to retrieve attention coefficients $\alpha \in [0,1]$ and upsample them back to the original dimension of $x$. The output of the attention gate is the element-wise product of attention coefficients and original input $x$, given by $\hat{x}$. Through attention gating, we filter out the noise while retaining global patterns of spatial variations.  

\subsection{U-Net for Spatial Interference}
Our proposed potential outcome prediction model is based on a U-Net architecture~\cite{ronneberger2015u}. It comprises three modules: downsampling blocks, upsampling blocks and a bottleneck block that acts as a bridge between the two. What distinguishes a U-Net architecture from a transformer based model is the use of skip connections between different upsampling and downsampling layers. In case of STCINet, these skip connections help retain the causal context in data.

\textbf{Downsample Block.} Our STCINet comprises two downsampling blocks. Each block consists of two 2D CNN layers, followed by a spatial attention layer, batch normalization layer and a 2D max pooling layer. The second block follows the same architecture with the difference of input shape. In every successive layer of the downsampler, we increment the output channels by a multiplicative factor of 2, as shown in Figure~\ref{fig:arch}. All CNN layers use the same $3\times 3$ kernel size filters. The ReLU activation function is used in all the downsampling layers. This part of our model helps learn low-level spatiotemporal dependencies in the data and identifies patterns needed for predicting spatial maps.

\textbf{Upsample Block.} This block learns from the low-level (downsampled) features and helps reconstruct the spatial map in the same dimension as the input but at a future timestep. Similar to the downsampler, the upsampler comprises two upsampling blocks. Every block comprises a $2\times2$ upsampling layer using the nearest interpolation method and a $2\times 2$ kernel size filter. Just before the the skip connection is built, we concatenate the output of each upsampling block with the feature map generated by the corresponding  downsampling block and pass it to the attention gate, as shown in Figure~\ref{fig:arch}. The output from attention gate is further concatenated with the output from the previous layer and finally passed through three 2D CNN layers. The output channel size of every CNN layer is reduced by a factor of 2 in order to regain the initial input dimension. Finally, a $1\times 1$ convolution with linear activation is applied to the upsampler's output to generate the predicted spatial map. 

\textbf{Custom Weighted Loss.} To jointly train the LFM module with the U-Net module, we customize the overall objective function to be a weighted sum of the two losses from LFM and U-Net. We further multiply the outcome of this custom loss with a $N \times M$ weight map $\mathbb{W}$. This element-wise multiplication is done to give treated units higher weightage than non-treated units. The purpose of subregion weighting is to help the model focus on treated areas irrespective of their overall spread. Our custom-loss function is given in Equation~\ref{eq:loss}:
\begin{equation}
    L_{tot} = \mathbb{W} \odot (\lambda_1 L_{lfm} + \lambda_2 L_{unet})
    \label{eq:loss}
\end{equation}
\begin{equation*}
  \text{where,    }  L_{lfm} = \frac{\Sigma_{\mathbb{N}}{({(X,Z)-\hat{\phi}})^2}}{|\mathbb{N}|}, L_{unet} = \frac{\Sigma_{\mathbb{N}}{({Y-\hat{Y}})^2}}{|\mathbb{N}|}
    %\label{eq:loss-lfm}
\end{equation*}
%\begin{equation}
%    L_{unet} = \frac{\Sigma_{|\mathbb{N}|}{({Y-\hat{Y}})^2}}{|\mathbb{N}|}
%    \label{eq:loss-unet}
%\end{equation}
$\lambda_1$ and $\lambda_2$ are hyperparameters to give weightage to the losses from LFM and U-Net, and $\lambda_1+\lambda_2 = 1$. 

\textbf{Treatment Effect Estimation.} Once the STCINet model is trained using custom loss, the trained model is used to make factual and counterfactual predictions by feeding observed and intervened treatment values to the model. The corresponding outcome predictions are used for estimating direct, indirect and lagged averaged treatment effects using the metrics defined in Section~\ref{definitions}.

\section{Experiments}
\label{sec4}
\subsection{Synthetic Dataset}
To test our method for tracking information flow in spatiotemporal data, we generate two variants of synthetic datasets to mimic a dominant physical process found in many geo-science applications, that is, diffusion. Diffusion is a physical process that describes the movement of particles or substances from regions of higher concentration to regions of lower concentration. Following this concept, we generate three spatiotemporal variables $X, Y$ and $Z$. Where $Z$ is an independent variable with spatial and temporal autocorrelations. $X$ is dependent on $Z$ and $Y$ is dependent on both $X$ and $Z$. The synthetic data is generated in Python using NumPy and SciPy libraries. The detailed description of our data generation process is given at GitHub\footnote{https://tinyurl.com/stcinet}. 
We utilize the causal coefficients $\alpha$, $\beta$, and $\gamma$ to incorporate causal influence in these variables. All diffusion coefficients ($D_x$, $D_y$, $D_z$) are set to 0.01.  We keep the temporal lag as 1 for all temporal dependencies, whereas the time step size ($dt$) is 0.1.

To model the spatial diffusion of each variable in the dataset, we perform a Laplacian operation $\nabla^2$ on each of them. Further, we employ a time-stepping loop to iteratively update the variables over multiple time steps. At each time step, the Laplacian of $X$, $Y$, and $Z$ is computed to model diffusion. The variables are then updated using their respective diffusion equations, incorporating time lags and dependencies between variables. For each time step $t$ from 1 to $T$, we update the variables using following equations: 
    
    \begin{equation}
    Z^t_{i,j} = Z^{t-1}_{i,j}  + dt \times \left( D_z \times \nabla^2 Z \right)
    \end{equation}
    \begin{equation}
    X^t_{i,j} = X^{t-1}_{i,j} + dt \times \left( D_x \times \nabla^2 X + \alpha \times \nabla^2 Z^{t-1}_{i,j} \right)
    \end{equation}
    \begin{equation}
    Y^t_{i,j} = Y^{t-1}_{i,j} + dt \times \left( D_y \times \nabla^2 Y +
    \beta \times \nabla^2 Z^{t-1}_{i,j} + \gamma \times \nabla^2 X^{t-1}_{i,j} \right)
    \end{equation}

We intervene on the treatment variable $X$ by applying an $update\_factor = 0.6$ to a specific sub-region $i=[10:15]$, $j=[10:15]$ of $X$ at time step $t$ to create the intervened scenario.
%    \begin{equation}
%    \hat{X}^{t}_{i,j} = update\_factor \times X^{t}_{i,j}
%    \end{equation}
The corresponding counterfactual outcome values $\hat{Y}$ are then generated by:
    \begin{equation}
        \hat{Y}^{t}_{i,j} = \hat{Y}^{t-1}_{i,j} + dt \times (D_y \times \nabla^2 \hat{Y} +
        \beta \times \nabla^2 Z^{t-1}_{i,j} +
        \gamma \times \nabla^2 \hat{X}^{t-1}_{i,j})
    \end{equation}

To incorporate a spillover effect of $X$ on the untreated regions, we add the mean of the per-pixel neighborhood of $X$ (excluding the pixel itself) in $Y$. A visualization of our synthetically generated data at different timesteps is given in Figure~\ref{fig:synth-data}.

\begin{figure}[ht!]
\centering
  \includegraphics[width=\textwidth]{synth_data.png}
  \caption{Potential outcome variable $Y$ at timesteps 10, 100, 1000, 2000 and 4000. Top row: Outcome under no intervention at different timesteps. Middle row: Intervened outcome at different timesteps. Bottom row: Spillover effects at different timesteps, which is the difference in intervened outcomes with and without the spatial interference.}
  \label{fig:synth-data}
\end{figure}
\subsection{Evaluation Metrics}
We provide the average treatment effect estimations for both synthetic and real-world datasets by reporting the Rooted Precision in Estimation of Heterogeneous Treatment (PEHE) scores based on the treatment effect metrics defined in Equations~\ref{eq:date}, \ref{eq:iate}, \ref{eq:late}. This metric is commonly used in machine learning literature for calculating the average error across the predicted ATEs~\cite{hill2011bayesian}. Additionally, we report the Root Mean Squared Error (RMSE) to evaluate the model's predictive performance. Both PEHE and RMSE can only be calculated for synthetic data which has ground truth information. 
Since it is a spatiotemporal 3D dataset, we customized the RMSE and PEHE metrics for our spatiotemporal models evaluation and report their respective error $\varepsilon$ using the following formulae:
%\begin{equation}
% \varepsilon_{RMSE} = \sqrt{\frac{\Sigma_{I}\Sigma_{J}%{\Big({Y[i,j]-\hat{Y}[i,j]}\Big)^2}}{N}}
%\label{eq:rmse}
%\end{equation}

\begin{equation*}
\varepsilon_{RMSE} = \sqrt{\frac{\Sigma_{I}\Sigma_{J}{\Big({Y[i,j]-\hat{Y}[i,j]}\Big)^2}}{|\mathbb{N}|}} 
 ,  \sqrt{\varepsilon_{PEHE}} = \sqrt{\frac{\Sigma_{I}\Sigma_{J}{\Big({\tau[i,j]-\hat{\tau}[i,j]}\Big)^2}}{|\mathbb{N}|}}
\label{eq:pehe}
\end{equation*}
where $\tau$ is the average treatment effect.
\subsection{Experimental Setup}
All our experiments are performed using the Amazon Web Services (AWS) cloud-based Elastic Compute Cloud (EC2) accelerated computing instances with high frequency 2.5 GHz (base) Intel Xeon Scalable Processor, 32 vCPUs and 64 GBs of GPU memory. Our STCINet model is trained using Keras Functional API with a Tensorflow backend and has around 293,000 trainable parameters. We trained our model using Adam optimizer with exponential decay of $e^{-0.1}$ in the learning rate after 10 epochs. The model was trained on 60 epochs with an early stopping criteria and batch size of 64 for all experiments. After hyperparameter tuning, we found the best performance with loss weightages as $\lambda_1 = 0.25$ and $\lambda_2 = 0.75$. The dataset is not split into training and test sets as our goal is to get outcome predictions on intervened treatment variables which automatically fulfills the unseen data requirement at test time.
\subsection{Ablation Study}
We test the performance of multiple variants of our proposed method on the synthetic datasets to identify the optimal configurations for spatiotemporal causal inference under spatial interference and temporal confounding. These variants and their corresponding ATE and PEHE scores are given in Table~\ref{tab:res-synth} for data with and without spatial interference. Here, $STCINet^\dagger$ refers to our predictive model without the $LFM$ or attention modules, $STCINet-NA$ refers to the spatiotemporal causal inference model with no attention (NA) mechanism, $STCINet-SA$ refers to the spatiotemporal causal inference model with spatial attention (SA) mechanism, $STCINet-AG$ refers to the spatiotemporal causal inference model with attention gating (AG) mechanism and $STCINet$ refers to the spatiotemporal causal inference model with $LFM$, spatial attention and attention gating mechanism. 

Observing the results on data with spatial interference in Table~\ref{tab:res-synth}, we see that $STCINet^\dagger$ yields lower $PEHE$ error than $STCINet$ for direct effect estimation (DATE). The exception of $STCINet^\dagger$ can be attributed to the fact that direct treatment effect is only estimated on the treated region where spillovers are easy to capture or non-existent, therefore we see the simplest variant performing the best on it. In case of no spatial interference, we observe that $STCINet$ gives the lowest (best) $PEHE$ error on direct, indirect and overall lagged treatment effects as compared to all its variants. It is also interesting to note that $STCINet-SA$ yields the second best results for IATE and LATE errors, in capturing treatment effects in the absence of spillover or interference, demonstrating the potential of spatial attention. We discuss the comparison with state-of-the-art (SOTA) methods in the next section.
%which proves the potential of our proposed model to work in both spatially varying and invariant settings.
%It is also interesting to note that $STCINet^\dagger$ yields the second best results for all DATE, IATE and LATE errors 

% Please add the following required packages to your document preamble:
% \usepackage{graphicx}
% \usepackage[normalem]{ulem}
% \useunder{\uline}{\ul}{}
% Please add the following required packages to your document preamble:
% \usepackage{graphicx}
\begin{table}[]
\centering
\caption{Related work comparison and ablation study of our proposed model on the synthetic data without and with spatial interference. \textbf{Bold} $\xrightarrow{}$ best results, \underline{underline} $\xrightarrow{}$ second best results.}
\label{tab:res-synth}
%\resizebox{\columnwidth}{!}{%
\begin{tabular}{|lcccc|}
\hline
\multicolumn{1}{|l|}{Model}      & \multicolumn{1}{l|}{$DATE(\sqrt{\varepsilon_{PEHE}})$}   & \multicolumn{1}{l|}{$IATE(\sqrt{\varepsilon_{PEHE}})$}   & \multicolumn{1}{l|}{$LATE(\sqrt{\varepsilon_{PEHE}})$}   & $\varepsilon_{RMSE}$   \\ \hline
\multicolumn{5}{|c|}{\textbf{Data without Spatial Interference}}                                                                    \\ \hline
\multicolumn{1}{|l|}{Deconfounder~\cite{bica2020estimating}} & \multicolumn{1}{l|}{1.8658}       & \multicolumn{1}{l|}{0.2008}       &    \multicolumn{1}{l|}{0.3890}   & 3.1450\\ \hline
\multicolumn{1}{|l|}{G-Net~\cite{li2021g}}        & \multicolumn{1}{l|}{1.3136} & \multicolumn{1}{l|}{0.0365} & \multicolumn{1}{l|}{0.0785}  & 2.0510\\ \hline
\multicolumn{1}{|l|}{Weather2Vec~\cite{tec2023weather2vec}}  & \multicolumn{1}{l|}{\textbf{1.1575}} & \multicolumn{1}{l|}{\textbf{0.0333}} & \multicolumn{1}{l|}{0.0784} & \underline{0.1640}\\ \hline
\multicolumn{1}{|l|}{STCINet$^\dagger$} & \multicolumn{1}{l|}{1.3129}          & \multicolumn{1}{l|}{0.0365}          & \multicolumn{1}{l|}{0.0785}          & 0.1660 \\ \hline
\multicolumn{1}{|l|}{STCINet-NA} & \multicolumn{1}{l|}{1.3120} & \multicolumn{1}{l|}{0.0363} & \multicolumn{1}{l|}{0.0783} & 0.1730 \\ \hline
\multicolumn{1}{|l|}{STCINet-SA} & \multicolumn{1}{l|}{1.2877} & \multicolumn{1}{l|}{\underline{0.0337}} & \multicolumn{1}{l|}{\underline{0.0752}} & 0.2200 \\ \hline
\multicolumn{1}{|l|}{STCINet-AG} & \multicolumn{1}{l|}{1.3140} & \multicolumn{1}{l|}{0.0364} & \multicolumn{1}{l|}{0.0785} & \textbf{0.1270} \\ \hline
\multicolumn{1}{|l|}{STCINet}           & \multicolumn{1}{l|}{\underline{1.2665}} & \multicolumn{1}{l|}{\underline{0.0337}} & \multicolumn{1}{l|}{\textbf{0.0744}} & 0.1690 \\ \hline
\multicolumn{5}{|c|}{\textbf{Data with Spatial Interference}}                                                                       \\ \hline
\multicolumn{1}{|l|}{Deconfounder~\cite{bica2020estimating}} & \multicolumn{1}{l|}{2.6580}       & \multicolumn{1}{l|}{0.2969}       &    \multicolumn{1}{l|}{0.5599}    & 6.4580 \\ \hline
\multicolumn{1}{|l|}{G-Net~\cite{li2021g}}        & \multicolumn{1}{l|}{1.4123} & \multicolumn{1}{l|}{0.0382} & \multicolumn{1}{l|}{0.0940}  & 6.7750 \\ \hline
\multicolumn{1}{|l|}{Weather2Vec~\cite{tec2023weather2vec}}  & \multicolumn{1}{l|}{\textbf{0.5103}} & \multicolumn{1}{l|}{0.0606} & \multicolumn{1}{l|}{0.0863}  & 0.3320\\ \hline
\multicolumn{1}{|l|}{STCINet$^\dagger$} & \multicolumn{1}{l|}{\underline{1.1959}} & \multicolumn{1}{l|}{\underline{0.0311}}          & \multicolumn{1}{l|}{\underline{0.0693}}          & 0.2930 \\ \hline
\multicolumn{1}{|l|}{STCINet-NA} & \multicolumn{1}{l|}{1.6182} & \multicolumn{1}{l|}{0.0454} & \multicolumn{1}{l|}{0.0971} & \textbf{0.2390} \\ \hline
\multicolumn{1}{|l|}{STCINet-SA} & \multicolumn{1}{l|}{1.6178} & \multicolumn{1}{l|}{0.0448} & \multicolumn{1}{l|}{0.0970} & 1.2790 \\ \hline
\multicolumn{1}{|l|}{STCINet-AG} & \multicolumn{1}{l|}{1.6179} & \multicolumn{1}{l|}{0.0453} & \multicolumn{1}{l|}{0.0972} & 0.6250 \\ \hline
\multicolumn{1}{|l|}{STCINet}           & \multicolumn{1}{l|}{1.5264}          & \multicolumn{1}{l|}{\textbf{0.0249}} & \multicolumn{1}{l|}{\textbf{0.0684}} & \underline{0.2790} \\ \hline
\end{tabular}%
%}
\end{table}

\subsection{Comparison with Baseline Methods}
Here, we compare STCINet with three state-of-the-art methods as our baselines, which are divided into the following two categories.

\textbf{Temporal deconfounding methods.} These methods perform causal inference on time-series data in the presence of time-varying confounders. We consider two such works, (i) Latent factor model modified for single treatment (Deconfounder)~\cite{bica2020estimating} and (ii) Deep learning based inverse propensity score method (G-Net)~\cite{li2021g} for calculating effects of time-varying continuous treatment. By reducing the spatial dimension, we apply these methods on our synthetic datasets.

\textbf{Spatial causal inference method.} Here, we consider a recent spatial causal inference method (Weather2vec) introduced for non-local spatial confounding. We consider the Weather2vec-AVG variant (see details in~\cite{tec2023weather2vec}) that can capture immediate neighborhood interference on the treated regions. To implement this spatial method, we reduce the temporal dimension by considering all data as independent samples irrespective of their temporal sequence. We present the results from these baselines in Table~\ref{tab:res-synth}. 

In case of direct treatment effect estimation (DATE) for both synthetic datasets, we notice Weather2Vec gives the lowest PEHE error which shows its potential of estimating interference effects on the treated region, however, the method fails to capture the interference on untreated region on data with spatial interference, quantified by indirect treatment effects (IATE). This shows the inability of Weather2Vec to capture spatial interference or spillover effects, further highlighting the significance of STCINet for overcoming the limitation of Weather2Vec. In case of IATE on data without spatial interference, we see a marginal difference in STCINet's and Weather2Vec results. Here it is important to note that this data does not possess spatial interference making it viable for Weather2Vec to have good estimations.
Overall, our model yields lowest PEHE error for both datasets in case of lagged average treatment effect (LATE), that accounts for temporal lag in effect estimation over the entire spatial region.
It is also interesting to see that G-Net's performance, despite being a time-series method, is comparable to our model, however, Deconfounder performs poorly in all scenarios yielding the worst performance overall.

\subsection{Case Study on Real-World Arctic Data}
\begin{figure*}[ht!]
\centering
\includegraphics[width=\textwidth]{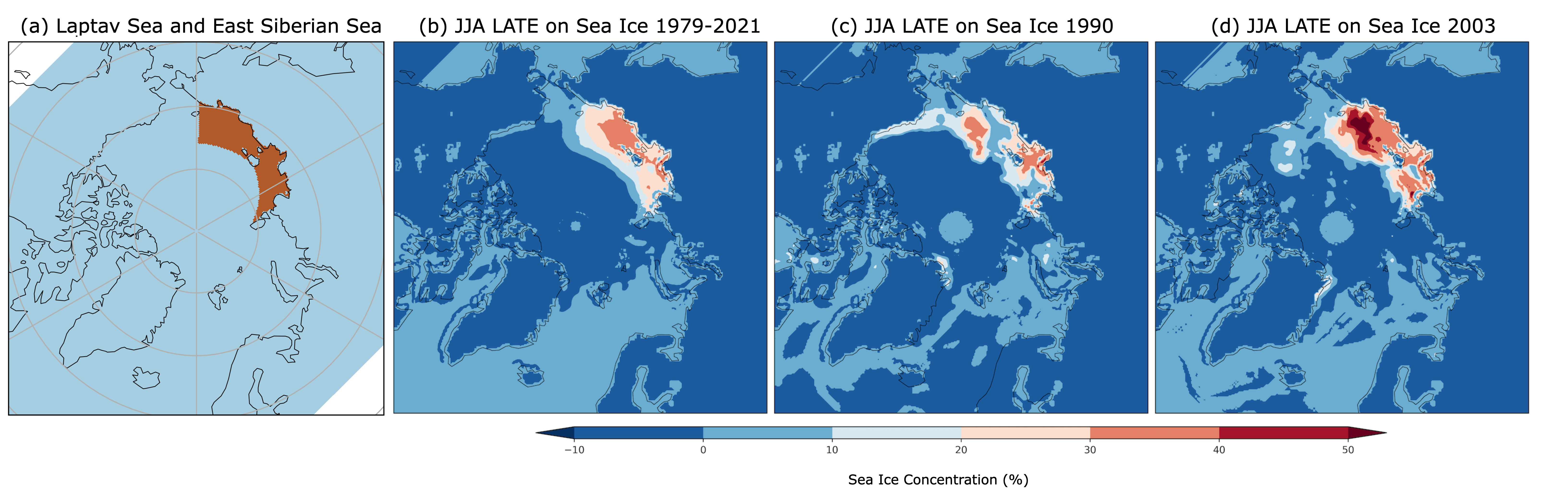}
  \caption{Case study on climate data when longwave radiations are reduced by 5\%. (a) Region of applying treatment, (b) Lagged Average treatment effect (LATE) on Summer SIC for 1979-2021, (c) LATE on Summer SIC for 1990, (d) LATE on Summer SIC for 2003.}
  \label{fig:arctic}
\end{figure*}

Here, we present a case study on real-world Arctic data where we estimate the causal influence of atmospheric processes on the Pan-Arctic sea ice concentrations (SIC), which have seen a continuous decline since 1979. This accelerated ice loss is prominently visible in Summers (JJA - June, July, August) where the minimum sea ice has reduced by more than 50\% of what it was in 1979~\cite{holland2019changing}. While identifying the true causes of ice melt is a complex task due to multiple thermodynamic feedbacks, a recent study suggests that one of the drivers of early melt in two Arctic sub-regions, namely East Siberian Sea and Laptav Sea, is the increase of downward longwave radiations (LWDN), with high ice melt observed in 1990 and 2003~\cite{huang2019survey}. Another study suggests that the sum of sensible and latent heat flux (HFX) plays an important role in Arctic's energy budget and has bidirectional causal links with LWDN and SIC~\cite{huang2021benchmarking}. Using STCINet and the data provided by~\cite{huang2021benchmarking}, we estimate the effect of these regional LWDN radiations on SIC on both regional and Pan-Arctic level at a lead time of one month. We include HFX, considering it a potential confounder in our study. The region of interest for applying treatment are the Laptav and East Siberian seas, as shown in Figure~\ref{fig:arctic}a. We  set $update\_factor$ (see details in Section 2.1) to be $-0.05$ for LWDN as our intervened treatment, which implies 5\% reduction in original LWDN values. Our trained STCINet model predicts an average of 4\% annual increase and a 44\% summer (JJA) increase in SIC in the Laptav and East Siberian seas if LWDN values were reduced by 5\% in that region. The direct treatment effect is visible in Figure~\ref{fig:arctic}b where we see a 42-year average difference (increase) in Laptav and East Siberian SIC. Our findings not only align with literature on the negative role of longwave radiations on sea ice melt, but also quantify the relations by estimating the direct and lagged average causal effects. More importantly, our model is able to capture the anomalous behavior in 1990 (Figure~\ref{fig:arctic}c) and 2003 (Figure~\ref{fig:arctic}d) where we observe that the effects of LWDN are not just restrained to the region of interest, but also spatially interfere with other Pan-Arctic regions influencing regional SIC~\cite{huang2019survey}. Through this case study, we demonstrate the potential of STCINet to provide insights into complex spatial and temporal relations of atmosphere and the ocean.  

\section{Related Work}
\label{sec5}
\label{related-work}
%Recently, causal inference has gained a lot of importance in economics, epidemiology and environmental science with a variety of statistical, regression-based and machine-learning methods proposed for estimating binary causal effects in i.i.d. and time-series data~\cite{yao2021survey}. 

The existing work in spatiotemporal causal inference is still foundational with much focus on the theoretical aspects. Wang et al. proposed causal inference framework for panel data with spatial and temporal interference under stable unit (SUTVA) assumption~\cite{wang2021causal}. Papadogeorgou et al. extended the potential outcome framework for point-process treatment and stochastic intervention~\cite{papadogeorgou2022causal}. Christiansen et al. proposed a non-parametric hypothesis test to develop causal models for multivariate spatiotemporal data~\cite{christiansen2022toward}. Owing to the intricate nature of spatiotemporal causal variations, the problem is often broken down to either time-series by fixing a region of interest, or spatial causal inference in time-invariant settings~\cite{runge2019inferring,ebert2014causal}. Here, we present the relevant literature in both domains. 

\textbf{Causal inference with Temporal Confounding.}
%Several methodologies have been proposed to estimate time-varying treatment effects~\cite{naimi2017introduction,menchetti2021estimating,google45950}. The works of~\cite{robins2000marginal} have introduced innovative approaches leveraging dynamic treatment effect models, dynamic structural equation modeling, and time-varying coefficient models, aiming to capture evolving treatment effects over time. However, these approaches often assume no confounding or struggle to address it adequately in time-series settings.
The challenge of confounding bias in time-series causal inference has received attention in recent literature. Notably, the techniques such as instrumental variables~\cite{CIV-2022}, propensity score matching~\cite{robins2000marginal}, and recurrent neural networks~\cite{moraffah2021causal,bica2020estimating} address confounding when estimating causal effects of time-invarying treatments. Few studies have explicitly addressed the joint challenge of estimating time-varying treatment effects while accounting for confounding bias in time-series data. Bica et al. attempted to bridge this gap, exploring methodologies that integrate time-series analysis with causal inference frameworks to disentangle multi-treatment effects from confounders in dynamic systems~\cite{bica2020estimating}. Their work struggles with single-cause treatment effect estimation. Recently, G-Net was proposed to tackle confounding of time-varying treatment using Long Short Term Memory (LSTM) model~\cite{li2021g}. These methods majorly work on binary treatments and there remains a notable gap in methodologies capable of effectively addressing both time-varying treatment effects and confounding bias in time-series data due to the strong ignorability condition.

\textbf{Causal inference for Spatial Interference.}
Most of the existing methods for spatial causal inference are basically spatial statistical techniques to study the interactions between spatial units in the presence of spatial confounding, spatial interference, or both~\cite{akbari2023spatial,reich2021review}. For instance,  Graham et al. used Poisson regression with spatial predictors to model spatial confounding and interference, but their approach does not focus on quantifying causal relations in data~\cite{graham2013quantifying}. Reich et al. and Giffin et al. both explored the utilization of spatial structure and generalized propensity scores to accommodate unmeasured confounding and interference~\cite{reich2021review,Giffin2022Generalized}. Wang et al. introduced a design-oriented framework for spatial experiments involving interference~~\cite{wang2020design}, while they later extended this to encompass spillover effects in panel data~\cite{wang2021causal}. In fact, Di et al. first introduced a spatial hierarchical Difference-in-Differences model for policy evaluation~\cite{di2016policy}, which Wang et al. delved further into design-based inference for spatial experiments considering unknown interference~\cite{wang2020design}. Most recently, Papadogeorgou et al. suggested a parametric method that concurrently tackles interference and bias arising from local and neighborhood unmeasured spatial confounding~\cite{papadogeorgou2023spatial}. These spatial methods are majorly an extension of difference-in-difference technique, which is inapplicable in time-varying domain, or propensity score methods which are computationally expensive methods and infeasible for continuous treatment effects estimation. 

Overall, there are several limitations of existing methods making them infeasible to offer generalized solutions for spatiotemporal data.  Some of the limitations include: (i) limiting work to specific applications, for instance, point process or panel data, (ii) mistaking spatial confounding with spatial interference, (iii) limiting scope to binary treatments, and (iv) inability to handle continuous or time-varying treatment assignment on spatial data. Our proposed STCINet model overcomes these limitations offering a state-of-the-art technique to perform causal inference on spatiotemporal data.

\section{Conclusion}
\label{sec6}
In this paper, we presented our deep learning based potential outcome model for spatiotemporal causal inference. We utilized latent factor modeling to reduce the bias due to time-varying confounding while leveraging the power of U-Net architecture and attention mechanism for capturing global and local spatial interference in data over time. Through empirical study on two synthetic datasets, we compared our method with state-of-the-art spatial and temporal inference methods to quantify direct (DATE), indirect (IATE), and lagged effects (LATE) on spatiotemporal data. We further provided a case study on real-world climate dataset and demonstrated the effectiveness of our proposed approach on quantifying the direct (sub-regional) and indirect (regional) effects of longwave radiations on sea ice concentration, paving paths for atmospheric scientists to adopt data driven methods to unravel important climate patterns. In the future, we would extend our work to estimate spatiotemporal causal inference in the presence of latent and spatially varying confounders. 

\section*{Acknowledgement}
This work is supported by NSF grants: CAREER: Big Data Climate Causality (OAC-1942714) and HDR Institute: HARP - Harnessing Data and Model Revolution in the Polar Regions (OAC-2118285).
\bibliographystyle{splncs04}
\bibliography{sample-base}
%
% ---- Bibliography ----
%
% BibTeX users should specify bibliography style 'splncs04'.
% References will then be sorted and formatted in the correct style.
%
% \bibliographystyle{splncs04}
% \bibliography{mybibliography}
%

\end{document}